\documentclass[10pt,twocolumn,letterpaper]{article}

\usepackage{iccv}
\usepackage{times}
\usepackage{epsfig}
\usepackage{graphicx}
\usepackage{amsmath}
\usepackage{amssymb}
\usepackage{booktabs}
\usepackage{mathtools}
\usepackage{comment}
\usepackage[linesnumbered,lined,ruled,noend]{algorithm2e}
\usepackage{diagbox}
\usepackage{multirow}
\usepackage{multicol}
\usepackage[numbers,sort]{natbib}
\usepackage{balance}
\usepackage{xcolor,soul}
\sethlcolor{lightgray}

\DeclarePairedDelimiter\norm{\lVert}{\rVert}%

\newtheorem{theorem}{Theorem}

\usepackage[pagebackref=true,breaklinks=true,letterpaper=true,colorlinks,bookmarks=false]{hyperref}

\iccvfinalcopy 

\def\iccvPaperID{1990} 

\ificcvfinal\fi

\begin{document}

\title{AdvRush: Searching for Adversarially Robust Neural Architectures}

\author{Jisoo Mok$^1$ ~~~~~~~ Byunggook Na$^1$ ~~~~~~~ Hyeokjun Choe$^1$  ~~~~~~~  Sungroh Yoon$^{1, 2, }$\thanks{Correspondence to: Sungroh Yoon $<$sryoon@snu.ac.kr$>$.}\\
$^1$ Department of Electrical and Computer Engineering, Seoul National University, Seoul, South Korea\\
$^2$ ASRI, INMC, ISRC, and Institute of Engineering Research, Seoul National University\\
{\tt\small \{magicshop1118, skqudrnrnabk, genesis1104, sryoon\}@snu.ac.kr}}

\maketitle
\ificcvfinal\thispagestyle{empty}\fi

\begin{abstract}
Deep neural networks continue to awe the world with their remarkable performance.
Their predictions, however, are prone to be corrupted by adversarial examples that are imperceptible to humans.
Current efforts to improve the robustness of neural networks against adversarial examples are focused on developing robust training methods, which update the weights of a neural network in a more robust direction.
In this work, we take a step beyond training of the weight parameters and consider the problem of designing an adversarially robust neural architecture with high intrinsic robustness.
We propose \textbf{AdvRush}, a novel adversarial robustness-aware neural architecture search algorithm, based upon a finding that independent of the training method, the intrinsic robustness of a neural network can be represented with the smoothness of its input loss landscape.
Through a regularizer that favors a candidate architecture with a smoother input loss landscape, AdvRush successfully discovers an adversarially robust neural architecture.
Along with a comprehensive theoretical motivation for AdvRush, we conduct an extensive amount of experiments to demonstrate the efficacy of AdvRush on various benchmark datasets.
Notably, on CIFAR-10, AdvRush achieves 55.91\% robust accuracy under FGSM attack after standard training and 50.04\% robust accuracy under AutoAttack after 7-step PGD adversarial training. 

\end{abstract}

\section{Introduction}
The rapid growth and integration of deep neural networks in every-day applications have led researchers to explore the susceptibility of their predictions to malicious external attacks.
Among proposed attack mechanisms, adversarial examples~\cite{szegedy2013intriguing}, in particular, raise serious security concerns because they can cause neural networks to make erroneous predictions with the slightest perturbations in the input data that are indistinguishable to the human eyes. 
This interesting property of adversarial examples has been drawing much attention from the deep learning community, and ever since their introduction, a plethora of defense methods have been proposed to improve the robustness of neural networks against adversarial examples~\cite{zhang2019theoretically, xie2019feature, lee2020adversarial}. 
However, there remains one important question that is yet to be explored extensively:
\textit{Can the adversarial robustness of a neural network be improved by utilizing an architecture with high intrinsic robustness?}
\textit{And if so, is it possible to automatically search for a robust neural architecture?}

\begin{figure}[t]
\begin{center}
   \includegraphics[width=\linewidth]{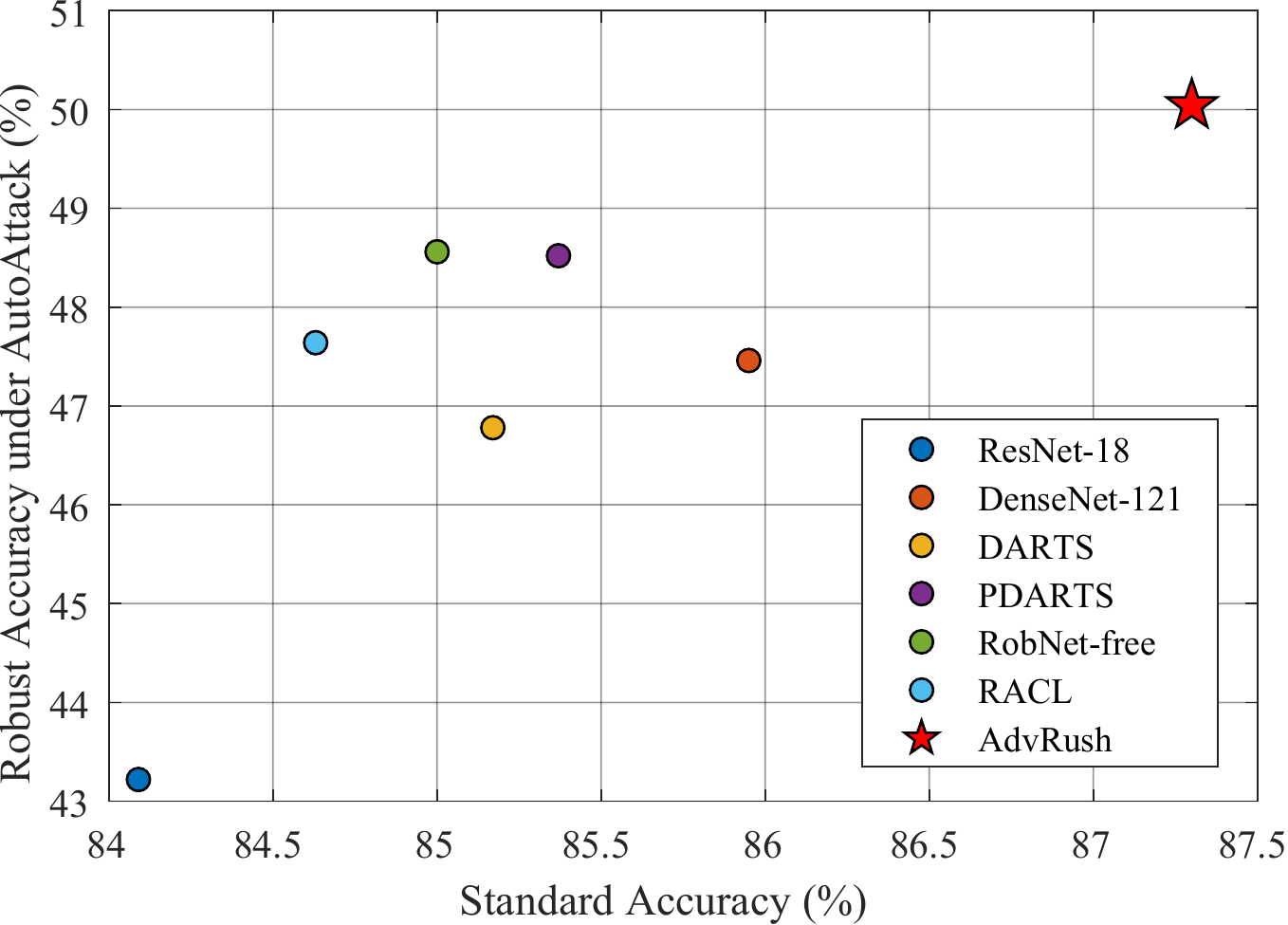}
\end{center}
\vspace{-7pt}
   \caption{Standard accuracy \textit{vs.} robust accuracy evaluation results on CIFAR-10 for various neural architectures and the neural architecture searched by AdvRush. All architectures are adversarially trained using 7-step PGD and evaluated under AutoAttack. AdvRush architecture achieves both standard accuracy-wise and robust accuracy-wise optimal frontiers.}
\label{fig:acc_rob_curve}
\vspace{-5pt}
\end{figure}

We tackle the problem of searching for a robust neural architecture by employing Neural Architecture Search (NAS)~\cite{le2017naswithRL}.
Due to the heuristic nature of designing a neural architecture, it used to take machine learning engineers with years of experience and expertise to fully exploit the power of neural networks.
NAS, a newly budding branch of automated machine learning, aims to automate this labor-intensive architecture search process.
As the neural architectures discovered automatically through NAS begin to outperform hand-crafted architectures across various domains~\cite{chen2019progressive, nekrasov2019fast, zhang2019customizable, tan2020efficientdet, mo2020neural}, more emphasis is being placed on the proper choice of an architecture to improve the performance of a neural network on a target task.

The primary objective of the existing NAS algorithms is concentrated on improving the standard accuracy, and thus, they do not consider the robustness of the searched architecture during the search process.
Consequently, they provide no guarantee of robustness for the searched architecture, since the ``no free lunch'' theorem for adversarial robustness prevents neural networks from obtaining sufficient robustness without additional effort~\cite{dohmatob2018limitations, bubeck2019adversarial, zhao2020bridging}.
In addition, the trade-off between the standard accuracy and the adversarial robustness indicates that maximizing the standard accuracy and the adversarial robustness cannot go hand in hand~\cite{tsipras2018robustness}, further necessitating a NAS algorithm designed specifically for adversarial robustness.

In this work, we propose a novel adversarial robustness-aware NAS algorithm, named \textbf{AdvRush}, which is a shorthand for ``\textbf{Adv}ersarially Robust Architecture \textbf{Rush}.''
AdvRush is inspired by a finding that the degree of curvature in the neural network's input loss landscape is highly correlated with intrinsic robustness, regardless of how its weights are trained~\cite{zhao2020bridging}.
Therefore, by favoring a candidate architecture with a smoother input loss landscape during the search phase, AdvRush discovers a neural architecture with high intrinsic robustness.
As shown in \figurename~\ref{fig:acc_rob_curve}, after undergoing an identical adversarial training procedure, the searched architecture of AdvRush simultaneously achieves the best standard and robust accuracies on CIFAR-10. 

We provide comprehensive experimental results to demonstrate that the architecture searched by AdvRush is indeed equipped with high intrinsic robustness.
On CIFAR-10, standard-trained AdvRush achieves 55.91\% robust accuracy (2.50\% improvement from PDARTS~\cite{chen2019progressive}) under FGSM, and adversarially-trained AdvRush achieves 50.04\% robust accuracy (3.04\% improvement from RobNet-free~\cite{guo2020meets}) under AutoAttack.
Furthermore, we evaluate the robust accuracy of AdvRush on CIFAR-100, SVHN, and Tiny-ImageNet to investigate its transferability; across all datasets, AdvRush consistently shows a substantial increase in the robust accuracy compared to other architectures.
For the sake of reproducibility, we have included the code and representative model files of AdvRush in the supplementary materials.

Our contributions can be summarized as follows:
\begin{itemize}
\item We propose AdvRush, a novel NAS algorithm for discovering a robust neural architecture. Because AdvRush does not require independent adversarial training of candidate architectures for evaluation, its search process is highly efficient.

\item  The effectiveness of AdvRush is demonstrated through comprehensive evaluation under a number of adversarial attacks. Furthermore, we validate the transferability of AdvRush to various benchmark datasets.

\item We provide extensive theoretical justification for AdvRush and complement it with the visual analysis of the discovered architecture. In addition, we provide a meaningful insight into what makes a neural architecture more robust against adversarial perturbations. 
\end{itemize}

\section{Related Works}
\subsection{Adversarial Attacks and Defenses}
Adversarial attack methods can be divided into white-box and black-box attacks. Under the white-box setting~\cite{papernot2016limitations, moosavi2016deepfool, kurakin2016adversarial, moosavi2017universal, xiao2018generating}, the attacker has full access to the target model, including its architecture and weights.   
FGSM~\cite{goodfellow2014explaining}, PGD~\cite{madry2017towards}, and CW~\cite{carlini2017towards} attacks are the famous white-box attacks, popularly used to evaluate the robustness of a neural network.
On the contrary, under the black-box setting, the attacker has limited to no access to the target model.
Thus, black-box attacks rely on a substitute model~\cite{papernot2017practical} or the target model's prediction score~\cite{chen2017zoo, brendel2017decision, su2019one, guo2019simple, ilyas2018black, moon2019parsimonious} to construct adversarial examples.

In response, numerous adversarial defense methods have been proposed to alleviate the vulnerability of neural networks to adversarial examples.
Adversarial training~\cite{goodfellow2014explaining} is known to be the most effective defense method.
By utilizing adversarial examples as training data, adversarial training plays a min-max game; while the inner maximum produces stronger adversarial examples to maximize the cross-entropy loss, the outer minimum updates the model parameters to minimize it. 
A large number of defenses now adopt some form of regularization or adversarial training to improve robustness~\cite{kannan2018adversarial, miyato2018virtual, ross2017improving, zhang2019defense, zhang2019theoretically, moosavi2019robustness, qin2019adversarial}.
Our work is closely related to the defense approaches that utilize a regularization term derived from the curvature information of the neural network's loss landscape to mimic the effect of adversarial training~\cite{moosavi2019robustness, qin2019adversarial}.

Apart from adversarial training, feature denoising~\cite{metzen2017detecting, meng2017magnet, samangouei2018defense, xie2019feature, xu2017feature} has also been proven to be effective at improving the robustness of a neural network.
Although gradient masking~\cite{song2017pixeldefend, papernot2017extending} may appear to be a viable defense method, it can easily be circumvented by attacks based on approximate gradients~\cite{athalye2018obfuscated, croce2020reliable}.

\subsection{Neural Architecture Search}
Early NAS algorithms based on evolutionary algorithm (EA)~\cite{real2017large, real2019regularized} or reinforcement learning (RL)~\cite{le2017naswithRL, zoph2018nasnet, raskar2016metaqnn} often required thousands of GPU hours to search for an architecture, making their immediate application difficult.
The majority of the computational overhead in these algorithms was caused by the need to train each candidate architecture to convergence and evaluate it. 
By using performance approximation techniques, recent NAS works were able to significantly expedite the architecture search process.
Examples of commonly-adopted performance approximation techniques include cell-based micro search space design~\cite{zoph2018nasnet} and parameter sharing~\cite{dean2018enas}.
Modern gradient-based algorithms that exploit such performance approximation techniques can be categorized largely into sampling-based NAS~\cite{xie2018snas, le2018oneshot, dong2019searching, zhang2020overcoming}, and continuous relaxation-based NAS~\cite{yang2019darts, chu2019fairdarts, chen2019progressive, chen2020stabilizing, xu2019pc} according to their candidate architecture evaluation method.

Following the successful acceleration of NAS, its application has become prevalent in various domains.
In computer vision, neural architectures discovered by modern NAS algorithms continue to produce impressive results in a variety of applications: image classification~\cite{chu2019fairdarts, chen2019progressive}, object detection~\cite{tan2020efficientdet, chen2019detnas, jiang2020sp, guo2020hit}, and semantic segmentation~\cite{liu2019auto, nekrasov2019fast, zhang2019customizable}.
NAS is also being applied to the domains outside computer vision, such as natural language processing~\cite{jiang2019improved} and speech processing~\cite{mo2020neural, kim2020evolved, qu2020evolutionary}.

Despite the proliferation of NAS research, only a limited amount of literature pertaining to the subject of robust neural architecture exists.
RobNet~\cite{guo2020meets} is the first work to empirically reveal the existence of robust neural architectures.
They randomly sample architectures from a search space and adversarially train each one of them to evaluate their robustness.
Because their method results in a huge computational burden, RobNet uses a narrow search space with only three possible operations.
RAS~\cite{kotyan2020towards}, an EA-based method, uses adversarial examples from a separate victim model to measure the robustness of candidate architectures, but their approach and objective are restricted to improving the robustness under black-box attacks. 
RACL~\cite{dong2020adversarially}, a gradient-based method, suggests to use the Lipschitz characteristics of the architecture parameters to achieve the target Lipschitz constant. 

\section{Theoretical Motivation}
Analyzing the topological characteristics of a neural network's loss landscape is an important tool for understanding its defining properties.
This section introduces theoretical backgrounds on the relationship between the loss landscape characteristics and adversarial robustness that inspired our method.
Section~\ref{3.1} provides definitions for parameter and input loss landscapes.
Using the provided definitions, section~\ref{3.2} shows how the degree of curvature in the input loss landscape of a neural architecture relates to its intrinsic adversarial robustness.

\subsection{Parameter and Input Loss Landscapes}\label{3.1}
We define a neural network as a function $f_A(\omega)$ where $f_A(\cdot)$ is its architecture, and $\omega$ is the set of trainable weight parameters.
Then, a loss function of a neural network can be expressed as $\mathcal{L}(f_A(\omega), x)$, where $x$ is the input data.

Since both $\omega$ and $x$ lie on a high-dimensional space, a direct visual analysis of $\mathcal{L}(f_A(\omega), x)$  is impossible.
Therefore, the high-dimensional loss surface of $\mathcal{L}(f_A(\omega), x)$ is projected onto an arbitrary low-dimensional space, namely a 2-dimensional hyperplane~\cite{goodfellow2014qualitatively, li2018visualizing}. 
Given two normalized projection vectors $e_x$ and $e_y$ of the 2-dimensional hyperplane and a starting point $o$, the points around $o$ are interpolated as follows:
\begin{equation}
    g(o, i, j, e_x, e_y) = o + ie_x + je_y.
\end{equation}
where $i$ and $j$ are the degrees of perturbations in the $e_x$ and $e_y$ directions, respectively.

Depending on the choice of the starting point $o$, the loss landscape can be visualized in either the parameter space ($\omega$) or the input space ($x$).
The computation of loss values for $o = \omega$ is formulated as: $\mathcal{L}(f_A(\omega + ie_x + je_y), x)$, which corresponds to the parameter loss landscape.
Similarly, for $o = x$, the loss values are computed as follows: $\mathcal{L}(f_A(\omega), x + ie_x + je_y)$, which corresponds to the input loss landscape.
In this paper, we will primarily focus on the input loss landscape. 

\subsection{Intrinsic Robustness of a Neural Architecture}\label{3.2}

Consider two types of loss functions for updating $\omega$ of an arbitrary neural network $f_A(\omega)$: a standard loss and an adversarial loss.
On one hand, standard training uses clean input data $x_{\mathrm{std}}$ to update $\omega$, such that the standard loss $\mathcal{L}_{\mathrm{std}} = \mathcal{L}(f_A(\omega), x_{\mathrm{std}})$ is minimized.
On the other hand, adversarial training uses adversarially perturbed input data $x_{\mathrm{adv}}$ to update $\omega$, such that the adversarial loss $\mathcal{L}_{\mathrm{adv}} = \mathcal{L}(f_A(\omega), x_{\mathrm{adv}})$ is minimized.
From here on, we refer to $f_A(\omega)$ after standard training and after adversarial training as $f_A(\omega_\mathrm{std})$ and as $f_A(\omega_\mathrm{adv})$, respectively.

Searching for a robust neural architecture is equivalent to finding an architecture $f_A(\cdot)$ with small $\max \mathcal{L}_\mathrm{adv}$, regardless of the training method.
Interestingly enough, the degree of curvature in the input loss landscape of $f_A(\cdot)$ is highly correlated with $\max \mathcal{L}_\mathrm{adv}$.
One way of quantifying the degree of curvature in the input loss landscape is through the eigenspectrum of $H$, the Hessian matrix of the loss computed with respect to input data.
From here on, we use the largest eigenvalue of the Hessian matrix $H$ to quantify the degree of curvature in the input loss landscape under second-order approximation and denote it as  $\lambda_{\mathrm{max}}(H)$~\cite{yu2019interpreting}.

Consider $f_A(\omega_\mathrm{std})$ and $f_A(\omega_\mathrm{adv})$, two independently trained neural networks with an identical neural architecture $f_A(\cdot)$.
We define $\Omega$ to be a set of $\omega_t$ which interpolates between $\omega_\mathrm{std}$ and $\omega_\mathrm{adv}$ along some parametric curve.
For instance, a quadratic Bezier curve~\cite{garipov2018loss} with endpoints $\omega_\mathrm{std}$ and $\omega_\mathrm{adv}$, connected by $\Omega$, can be expressed as follows:
\begin{equation}
\begin{split}
    \phi_\Omega(p) & = (1-p)^2 \omega_\mathrm{std} + 2p(1-p)\Omega + p^2 \omega_\mathrm{adv}, \\
    & ~~~~~ \textnormal{where}~0 \leq p \leq 1\textnormal{,~and}~\Omega = \{\omega_t\}_{t=1}^T. 
\end{split}
\end{equation}
$T$ denotes the number of $\omega_t$ (\ie., the number of bends in the curve).
For every $f_A(\omega_t)$, a high correlation between its $\lambda_\mathrm{max}(H_t)$ and $\max \mathcal{L}_\mathrm{adv}$ are observed~\cite{zhao2020bridging}.
The following theorem provides theoretical evidence for the empirically observed correlation between the two.

\begin{theorem} \textnormal{(Zhao} et al.~\textnormal{\cite{zhao2020bridging})}
Consider the maximum adversarial loss $\max_{\norm{\delta} \leq \epsilon} \mathcal{L}(f_A(\omega_t), x_\mathrm{std}+\delta)$ of any $f_A(\omega_t)$ on the path $\phi_\Omega(p)$, where $x_\mathrm{std} + \delta$ represents $x_\mathrm{adv}$ with $\norm{\delta}$ confined by an $\epsilon$-ball. Assume: \\
(a) the standard loss $\mathcal{L}_\mathrm{std} = \mathcal{L}(f_A(\omega_t), x_\mathrm{std})$ on the path is a constant for all $f_A(\omega_t)$. \\
(b) $\mathcal{L}(f_A(\omega_t) , x_\mathrm{std} + \delta) \approx \mathcal{L}(f_A(\omega_t), x_\mathrm{std}) + (g_t)^{T}\delta + \frac{1}{2}\delta^{T}H_t\delta$ for small $\delta$, where $g_t$ and $H_t$ denote the gradient and the Hessian of $\mathcal{L}_t$ at clean input $x_{\mathrm{std}}$.
Let $c$ denote the normalized inner product in absolute value for the largest eigenvector $v$ of $H_t$  and $g_t, \frac{(g_t)^{T}v}{\norm{g_t}} = c$. Then, we have \\
\begin{equation}\label{correlation}
    \max_{\norm{\delta} \leq \epsilon}~\mathcal{L}(f_A(\omega_t), x_\mathrm{std} + \delta) \sim \lambda_{\mathrm{max}}(H_t)~as~c \rightarrow 1. 
\end{equation} 
\end{theorem}
Please refer to Zhao \textit{et al.}~\cite{zhao2020bridging} for the proof. The left-hand side of Eq.~(\ref{correlation}) corresponds to $\max \mathcal{L}_\mathrm{adv}$ of all $f_A(\omega_t)$.
For $f_A(\omega_\mathrm{std})$ specifically, Eq.~(\ref{correlation}) can be re-written as follows:
\begin{equation}\label{thm2_eq}
    \max_{\norm{\delta} \leq \epsilon}~\mathcal{L}(f_A(\omega_\mathrm{std}), x_\mathrm{std} + \delta) \sim \lambda_{\mathrm{max}}(H_\mathrm{std})~as~c \rightarrow 1.
\end{equation}

Geometrically speaking, adversarial attack methods perturb $x_{\mathrm{std}}$ in a direction that maximizes the change in $\mathcal{L}_{\mathrm{std}}$.
The resulting adversarial examples $x_{\mathrm{adv}}$ fool $f_A(w_\mathrm{t})$ by targeting the steep trajectories on the input loss landscape, crossing the decision boundary of a neural network with as little effort as possible~\cite{yu2019interpreting}.
Therefore, the more curved the input loss landscape of $f_A(\omega_\mathrm{t})$ is, its predictions are more likely to be corrupted by adversarial examples.

\section{Methodology}
Based on the findings in Section~\ref{3.2}, the problem of searching for an adversarially robust neural architecture can be re-formulated into the problem of searching for a neural architecture with a smooth input loss landscape. 
Since it is computationally infeasible to calculate the curvature of $f_A(\omega_\mathrm{t})$ for every $\omega_\mathrm{t}$, we opt to evaluate candidate architectures under after standard training:
\begin{equation}\label{general_framework}
\begin{aligned}
    f^*_A(\cdot) 
    & = \mathrm{argmin}_{f_A(\cdot)} \max_{\norm{\delta} \leq \epsilon} \mathcal{L}(f_A(\omega_\mathrm{std}), x+\delta) \\
    & = \mathrm{argmin}_{f_A(\cdot)} \lambda_\mathrm{max}(H_\mathrm{std}),
\end{aligned}
\end{equation}
where $H_\mathrm{std}$ refers to the Hessian of $\mathcal{L}_\mathrm{std}$ of $f_A(\omega_\mathrm{std})$ at clean input $x_\mathrm{std}$, and $\lambda_\mathrm{max}(H_\mathrm{std})$ refers to the largest eigenvalue of $H_\mathrm{std}$.

Therefore, during the search process, AdvRush penalizes candidate architectures with large $\lambda_\mathrm{max}(H_\mathrm{std})$.
By favoring a candidate neural architecture with a smoother loss landscape, AdvRush effectively searches for a robust neural architecture. 
In this section, we show how the objective of AdvRush in Eq.~(\ref{general_framework}) can be incorporated into the bi-level optimization problem of NAS~\cite{yang2019darts} and provide a mathematical derivation for approximating the Hessian matrix.

\subsection{AdvRush Framework}\label{4.1}

Standard training of each candidate architecture and evaluating its robustness against adversarial examples incur tremendous computational overhead to derive $f_A^*(\cdot)$ in Eq.~(\ref{general_framework}). 
Thus, AdvRush employs differentiable architecture search~\cite{yang2019darts} to allow for a simultaneous evaluation of all candidate architectures.
AdvRush starts by constructing a weight-sharing supernet~\cite{yang2019darts}, $f_\mathrm{super}(\cdot)$, from which candidate architectures inherit weight parameters $\omega$.
Following the convention in differentiable architecture search~\cite{yang2019darts}, we represent the supernet in the form of a directed acyclic graph (DAG) with $N$ number of nodes.
Each node $\{x^{(i)}\}_{i=1}^N$ of this DAG corresponds to a feature map, and each edge $(i, j)$ corresponds to a candidate operation $o^{(i,j)}$ that transforms $x^{(i)}$.
Each intermediate node of the graph is computed based on all of its predecessors:
\vspace{-0.25em}
\begin{equation}
x^{(j)} = \sum_{i < j} o^{(i,j)}(x^{(i)}).
\vspace{-0.5em}
\end{equation}

To make the search space continuous for gradient-based optimization, categorical choice of a particular operation is continuously relaxed by applying a softmax function over all the possible operations:
\vspace{-0.25em}
\begin{equation}
\bar{o}^{(i,j)}(x) = \sum_{o \in \mathcal{O}} \frac{\mathrm{exp}(\alpha_{o}^{(i,j)})}{\sum_{o' \in \mathcal{O}} \mathrm{exp}(\alpha_{o'}^{(i,j)})} o(x),
\vspace{-0.5em}
\end{equation}
where $\alpha^{(i,j)}$ is a set of operation mixing weights (\ie., architecture parameters).
$\mathcal{O}$ is the pre-defined set of operations that are used to construct the supernet.
By definition, the size of $\alpha^{(i,j)}$ must be equal to $|\mathcal{O}|$.

Through continuous relaxation, both the architecture parameters $\alpha$ and the weight parameters $\omega$ in the supernet can be updated via gradient descent.
Once the supernet converges, a single neural architecture can be obtained through the discretization step: $o^{(i,j)} = \mathrm{argmax}_{o \in \mathcal{O}} \alpha_o^{(i,j)}$.
The objective of AdvRush now becomes to update $\alpha$ to induce smoothness in the input loss landscape of the standard-trained $f_\mathrm{super}(\omega_\mathrm{std})$, such that the final discretization step will yield $f^*_A(\cdot)$ in Eq.~(\ref{general_framework}).

AdvRush accomplishes the above objective by driving the eigenvalues of $H_\mathrm{std}$ of $f_\mathrm{super}(\omega_\mathrm{std})$ to be small.
Consequently, their maximum, $\lambda_\mathrm{max}(H_\mathrm{std})$ will also be small.
Let $\lambda_1, ... , \lambda_d$ denote the eigenvalues of $H_\mathrm{std}$.
Our ideal loss term can be defined as the Frobenius norm of $H_\mathrm{std}$: $\mathcal{L}_\lambda = \sqrt{\sum_i (\lambda_i)^2} = \sqrt{\mathrm{Tr}(H_\mathrm{std}^{T}H_\mathrm{std})} = \norm{H_\mathrm{std}}_\mathrm{F}$.
The resulting bi-level optimization problem of AdvRush can be expressed as follows:
\begin{equation}
\begin{aligned}
\mathrm{min}_\alpha & ~\mathcal{L}(\omega_\mathrm{std}(\alpha), \alpha; \mathcal{D}_\mathrm{val}) + \gamma \mathcal{L}_\lambda, \\
    & \mathrm{where} ~\mathcal{L}_\lambda = \norm{H_\mathrm{std}}_\mathrm{F}, \\
& \mathrm{s.t.} ~\omega_\mathrm{std}(\alpha) = \mathrm{argmin}_\omega~\mathcal{L}(\omega, \alpha; \mathcal{D}_\mathrm{train}).
\end{aligned}
\end{equation} 
$\mathcal{D}_\mathrm{train}$ refers to training data, and $\mathcal{D}_\mathrm{val}$ to validation data. 
In the following section, we show how $\mathcal{L}_\lambda$ can be computed without  a significant increase in the search cost of AdvRush.

\subsection{Approximation of $\mathcal{L}_\lambda$}\label{4.2}

$\norm{H_\mathrm{std}}_\mathrm{F}$ can be expressed in terms of $l2$ norm: $\norm{H_\mathrm{std}}_\mathrm{F} = E[\norm{H_\mathrm{std}z}_2]$, where the expectation is taken over $z \sim N(0, I_d)$.
Because the direct computation of $H_\mathrm{std}$ is expensive, we linearly approximate it through the finite difference approximation of the Hessian:
\begin{equation} \label{approx_hess}
\begin{aligned}
H_\mathrm{std}z \approx \frac{l(x_\mathrm{std}+hz) - l(x_\mathrm{std})}{h}, \\
\mathrm{where} ~l(x) = \nabla_{x} \mathcal{L}(f_\mathrm{super}(\omega_\mathrm{std}), x),
\end{aligned}
\vspace{-0.5em}
\end{equation}
where $h$ controls the scale of the loss landscape on which we induce smoothness. 
However, computing multiple $H_\mathrm{std}z$ in directions drawn from $z \sim N(0, I_d)$ and taking its average would be computationally inefficient because each computation of $H_\mathrm{std}z$ requires calculation of the gradient.
Therefore, we minimize the input loss landscape along the the high curvature direction, $z^* = \frac{\mathrm{sign}(\nabla_{x}\mathcal{L}(x_{\mathrm{std}})}{\norm{\mathrm{sign}(\nabla_{x}\mathcal{L}(x_{\mathrm{std}})}}$ to maximize the effect of $\mathcal{L}_\lambda$~\cite{moosavi2019robustness, jetley2018friends, fawzi2018empirical}.

With the approximated $\mathcal{L}_\lambda$, the bi-level optimization problem of AdvRush can be expressed as:
\vspace{-0.2em}
\begin{equation}
\label{final_eq}
\begin{aligned}
\mathrm{min}_\alpha & ~\mathcal{L}(\omega_\mathrm{std}(\alpha), \alpha; \mathcal{D}_\mathrm{val}) + \gamma \mathcal{L}_\lambda, \\
    & \mathrm{where}~\mathcal{L}_\lambda = \norm{l(x_{\mathrm{val}}+hz^*) - l(x_{\mathrm{val}})} \\
    &~~~~~~~~~~~ \mathrm{and}~l(x) = \nabla_{x} \mathcal{L}(f_\mathrm{super}(\omega_\mathrm{std}), x), \\
& \mathrm{s.t.} ~\omega_\mathrm{std}(\alpha) = \mathrm{argmin}_\omega~\mathcal{L}(\omega, \alpha; \mathcal{D}_\mathrm{train}).
\end{aligned}
\vspace{-0.2em}
\end{equation} 
$x_\mathrm{val}$ is the clean input data from $\mathcal{D}_\mathrm{val}$.
The value of $h$ in the denominator of Eq.~(\ref{approx_hess}) is absorbed by the regularization strength $\gamma$.
The remaining $h$ in Eq.~(\ref{final_eq}) is treated as a hyperparameter of AdvRush.

Because the loss landscape of a randomly initialized supernet is void of useful information, in AdvRush, we warm up the architecture parameters $\alpha$ and the weight parameters $\omega$ of the supernet without $\mathcal{L}_\lambda$.
Upon the completion of the warm-up process, we introduce $\mathcal{L}_\lambda$ for additional epochs.
Please refer to the Appendix for the comprehensive pseudo code of AdvRush search process.

\section{Experiments}
In the following sections, we present extensive experimental results to demonstrate the effectiveness of AdvRush.
Notably, we observe that the neural architecture discovered by AdvRush consistently exhibits superior robust accuracy across various benchmark datasets: CIFAR-10~\cite{krizhevsky2009learning}, CIFAR-100~\cite{krizhevsky2009learning}, SVHN~\cite{netzer2011reading}, and Tiny-ImageNet~\cite{deng2009imagenet}.
NVIDIA V100 GPU and NVIDIA GeForce GTX 2080 Ti GPU are used for our experiments.

\subsection{Experimental Settings} \label{5.1}
\noindent\textbf{AdvRush} 
Following the convention in NAS literature~\cite{yang2019darts, chu2019fairdarts}, the CIFAR-10 dataset is used to execute AdvRush.
We use DARTS~\cite{yang2019darts} as the backbone differentiable architecture search algorithm because it is one of the most widely-benchmarked algorithms in NAS.
For the search phase, the training set of CIFAR-10 is evenly split into two: one for updating $\omega$ and the other for updating $\alpha$.
We generally follow the hyperparameter setting in DARTS~\cite{yang2019darts}, with a few modifications.
We run AdvRush for total of 60 epochs, 50 of which are allocated for the warm-up process.
The value of $h$ in Eq.~(\ref{final_eq}) is set to be 1.5.
We use a batch size of 32 for all epochs.
To update $\omega$, we use momentum SGD, with the initial learning rate of 0.025, momentum of 0.9, and weight decay factor of 3e-4.
To update $\alpha$, we use Adam with the initial learning rate of 3e-4, momentum of (0.5, 0.999), and weight decay factor of 1e-3.

\noindent\textbf{Standard \& Adversarial Training}
Based on the model size measured in the number of parameters, following hand-crafted and NAS-based architectures are used for comparison: ResNet-18~\cite{he2016deep}, DenseNet-121~\cite{huang2017densely}, DARTS~\cite{yang2019darts}, PDARTS~\cite{chen2019progressive}, RobNet-free~\cite{guo2020meets}, and RACL~\cite{dong2020adversarially}.
For a fair evaluation of each architecture's intrinsic robustness, all the tested architectures are trained using identical training settings.
1) Standard training: We train all architectures for 600 epochs. We use SGD optimizer with the initial learning rate of 0.025, which is annealed to zero through cosine scheduling. We set the weight decay factor to be $3 \times 10^{-4}$.
2) Adversarial training: We use 7-step PGD training~\cite{madry2017towards} with the step size of 0.01 and the total perturbation scale $\epsilon$ of 0.031 ($= 8/255$) to train all architectures.
In addition to the evaluation on CIFAR-10, we evaluate the transferability of AdvRush by adversarially training the searched architecture on the following datasets: CIFAR-100, SVHN, and Tiny-ImageNet.
Because a different set of hyperparameters is used for each dataset, the dataset configurations and the summary of hyperparameters for adversarial training are provided in the Appendix.
\begin{table*}[t]
\centering
\caption{Evaluation of robust accuracy on CIFAR-10 under white-box attacks. The best result in each column is in bold, and the second best result is underlined. PGD$^{20}$ and PGD$^{100}$ refer to PGD attack with 20- and 100-iterations, respectively. AA refers to the final evaluation result after completing the standard group of AutoAttack methods. All attacks are $l_{\infty}$-bounded with total perturbation scale of 0.031.}
\setlength{\tabcolsep}{5pt}
\vspace{5pt}
\renewcommand{\arraystretch}{1.13}
\begin{tabular}{l|r|cccccc|cc}
\toprule
& & \multicolumn{6}{c|}{Adversarially Trained} & \multicolumn{2}{c}{Standard Trained} \\ 
Model & Params & Clean & FGSM & PGD$^{20}$ & PGD$^{100}$ & APGD\textsubscript{CE} & AA & Clean & FGSM \\
\bottomrule \toprule
ResNet-18 & 11.2M & 84.09\% & 54.64\% & 45.86\% & 45.53\% & 44.54\% & 43.22\% & 95.84\% & 50.71\% \\
DenseNet-121 & 7.0M & \underline{85.95\%} & 58.46\% & 50.49\% & 49.92\% & 49.11\% & 47.46\% & 95.97\% & 45.51\% \\
\midrule
DARTS & 3.3M & 85.17\% & 58.74\% & 50.45\% & 49.28\% & 48.32\% & 46.79\% & 97.46\% & 50.56\% \\
PDARTS & 3.4M & 85.37\% & 59.12\% & 51.32\% & 50.91\% & 49.96\% & 48.52\% & \underline{97.49\%} & \underline{54.51\%} \\
RobNet-free & 5.6M & 85.00\% & \underline{59.22\%} & \underline{52.09\%} & \underline{51.14\%} & \underline{50.41\%} & \underline{48.56\%} & 96.40\% & 36.99\% \\
RACL & 3.6M & 84.63\% & 58.57\% & 50.62\% & 50.47\% & 49.42\% & 47.64\% & 96.76\% & 52.38\% \\
\midrule
AdvRush & 4.2M & \textbf{87.30\%} & \textbf{60.87\%} & \textbf{53.07\%} & \textbf{52.80\%} & \textbf{51.83\%} & \textbf{50.05\%} & \textbf{97.58\%} & \textbf{55.91\%} \\
\bottomrule
\end{tabular}
\vspace{-5pt}
\label{table:whitebox}
\end{table*}
\begin{table*}[t]
\centering
\caption{Evaluation of robust accuracy on CIFAR-10 under black-box attacks. Adversarial examples from the source model are generated with PGD$^{20}$. The best result in each row is in bold. The robust accuracy of each architecture under white-box attack is highlighted in gray.} 
\setlength{\tabcolsep}{5pt}
\vspace{5pt}
\renewcommand{\arraystretch}{0.9}
\begin{tabular}{l|ccccccc}
\toprule
\diagbox{Source}{Target} & ResNet-18 & DenseNet-121 & DARTS & PDARTS & RobNet-free & RACL & AdvRush \\
\midrule 
ResNet-18 & \hl{45.86\%} & 66.31\% & 66.46\% & 67.54\% & {67.89\%} & 66.20\% & \textbf{68.52\%} \\
\midrule
DenseNet-121 & 63.14\% & \hl{50.49\%} & 65.58\% & {66.60\%} & 66.58\% & 65.17\% & \textbf{67.18\%} \\
\midrule
DARTS & 64.40\% & 60.84\% & \hl{50.45\%} & 65.90\% & 65.54\% & 64.55\% & \textbf{66.89\%} \\
\midrule
PDARTS & 64.46\% & 60.44\% & 64.73\% & \hl{51.32\%} & 65.61\% & 64.23\% & \textbf{66.71\%} \\
\midrule
RobNet-free & 64.13\% & 61.03\% & 64.32\% & 65.30\% & \hl{52.09\%} & 63.72\% & \textbf{65.40\%} \\
\midrule
RACL & 64.49\% & 60.46\% & 64.67\% & 65.70\% & 65.73\% & \hl{50.62\%} & \textbf{66.58\%} \\
\midrule
AdvRush & 64.43\% & 60.98\% & 64.78\% & 65.24\% & 64.55\% & 64.23\% & \hl{53.07\%} \\
\bottomrule
\end{tabular}
\vspace{-8pt}
\label{table:blackbox}
\end{table*}

\subsection{White-box Attacks}
We evaluate the adversarial robustness of architectures standard and adversarially trained on CIFAR-10 using various white-box attacks.
Standard-trained architectures are evaluated under FGSM attack, while adversarially-trained architectures are evaluated under FGSM~\cite{goodfellow2014explaining}, PGD$^{20}$, PGD$^{100}$~\cite{madry2017towards}, and the standard group of AutoAttack (APGD\textsubscript{CE}, APGD\textsuperscript{T}, FAB\textsuperscript{T}, and Square)~\cite{croce2020reliable}.
White-box attack evaluation results are presented in \tablename~\ref{table:whitebox}.
After both training schemes, AdvRush achieves the highest standard and robust accuracies, indicating that the AdvRush architecture is in fact equipped with higher intrinsic robustness.
Even with the cell-based constraint, the robust accuracy of AdvRush under AutoAttack is higher than that of RobNet-free, which removes the cell-based constraint~\cite{guo2020meets}. 
In addition, the high robust accuracy of AdvRush after AutoAttack evaluation implies that the AdvRush architecture does not unfairly benefit from obfuscated gradients.

\subsection{Black-box Attacks}
To evaluate the robustness of the searched architecture under a black-box setting, we conduct transfer-based black-box attacks among adversarially-trained architectures.
Black-box evaluation results are presented in \tablename~\ref{table:blackbox}.
Clearly, regardless of the source model used to synthesize adversarial examples, AdvRush is most resilient against tranfer-based black-box attacks.
When considering each model pair, AdvRush generates stronger adversarial examples than its counterpart; for instance, AdvRush $\rightarrow$ DARTS achieves the attack success rate (\ie., 100\% - robust accuracy) of 35.22\%, while DARTS $\rightarrow$ AdvRush achieves the attack success rate of 33.11\%.

\subsection{Transferability to Other Datasets}
We transfer the architecture searched on CIFAR-10 to other datasets to evaluate its general applicability.
The standard and the robust accuracy evaluation results on CIFAR-100, SVHN, and Tiny-ImageNet are presented in~\tablename~\ref{table:datasets}, which shows that the AdvRush architecture is highly transferable. 
Notice that on SVHN dataset, AdvRush experiences an extremely low drop in accuracy under both FGSM and PGD$^{20}$ attacks.
This result may imply that on easier datasets, such as SVHN, having a robust architecture could be sufficient for achieving high robustness, even without advanced adversarial training methods.
Please refer to the Appendix for the full evaluation results.
\begin{table}[t]
\centering
\caption{Evaluation of robust accuracy on various datasets under white-box attacks. We transfer the AdvRush architecture searched on CIFAR-10 to CIFAR-100, SVHN, and Tiny-ImageNet and evaluate its robustness against FGSM and PGD$^{20}$ attacks. All attacks are $l_{\infty}$-bounded with total perturbation scale of 0.031.}
\setlength{\tabcolsep}{3pt}
\vspace{5pt}
\renewcommand{\arraystretch}{1.1}
\begin{tabular}{l|l|ccc}
\toprule
Dataset & Model & Clean & FGSM & PGD$^{20}$ \\
\bottomrule\toprule
\multirow{4}{*}{CIFAR-100} & ResNet-18 & 55.57\% & 26.03\% & 21.44\% \\
& DenseNet-121 & \textbf{62.33\%} & \underline{34.68\%} & \underline{28.67\%} \\
& PDARTS & 58.41\% & 30.35\% & 25.83\% \\
& AdvRush & \underline{58.73\%} & \textbf{39.51\%} & \textbf{30.15\%} \\
\midrule
\multirow{4}{*}{SVHN} & ResNet-18 & 92.06\% & 88.73\% & 69.51\% \\
& DenseNet-121 & 93.72\% & 91.78\% & 76.51\% \\
& PDARTS & \underline{95.10\%} & \underline{93.01\%} & \underline{89.58\%} \\
& AdvRush & \textbf{96.53\%} & \textbf{94.95\%} & \textbf{91.14\%} \\
\midrule
& ResNet-18 & 36.26\% & 16.08\% & 13.94\% \\
Tiny- & DenseNet-121 & \textbf{47.56\%} & 22.98\% & 18.06\% \\
ImageNet & PDARTS & \underline{45.94\%} & \underline{24.36\%} & \underline{22.74\%} \\
& AdvRush & 45.42\% & \textbf{25.20\%} & \textbf{23.58\%} \\
\bottomrule
\end{tabular}
\vspace{-0.5em}
\label{table:datasets}
\end{table}

\section{Discussion}
\subsection{Effect of Regularization Strength}
\begin{figure}
    \centering
    \includegraphics[width=\linewidth]{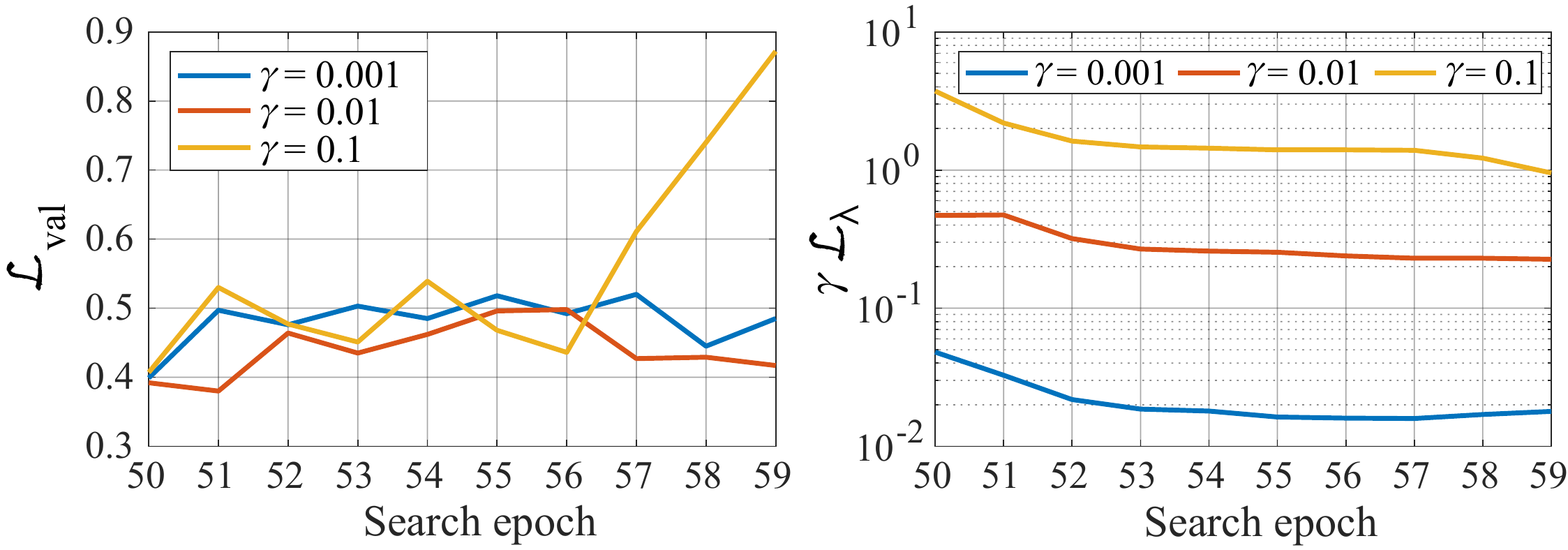}
    \caption{Search epoch \textit{vs.} $\mathcal{L}_\mathrm{val}$ (left) and $\gamma\mathcal{L}_\mathrm{\lambda}$ (right). $\mathcal{L}_\mathrm{val}$ is plotted in linear scale, and $\gamma\mathcal{L}_\mathrm{\lambda}$ is plotted in logarithmic scale. It is clear that large $\gamma\mathcal{L}_\mathrm{\lambda}$ causes explosion in $\mathcal{L}_\mathrm{val}$.}
    \label{fig:loss_search}
    \vspace{-3pt}
\end{figure}

The regularization strength $\gamma$ is empirically set to be 0.01 to match the scale of $\mathcal{L}_\mathrm{val}$ and $\mathcal{L}_\lambda$. 
The search results of other values of $\gamma$ in~\tablename~\ref{table:gamma}. 
Regardless of the change in $\gamma$, the robust accuracy of the searched architecture is higher than that of the other tested architectures (\tablename~\ref{table:whitebox}). 
For large $\gamma$, however, the searched architecture experiences a significant drop in standard accuracy.
Therefore, we conclude that AdvRush is not unduly sensitive to the tuning of $\gamma$, as long as it is sufficiently small.
We track the change in $\mathcal{L}_\mathrm{val}$ and $\gamma\mathcal{L}_\lambda$ for different values of $\gamma$ and plot the result in~\figurename~\ref{fig:loss_search}; clearly, large $\gamma\mathcal{L}_\lambda$ causes $\mathcal{L}_\mathrm{val}$ to explode, thereby disrupting the search process.
Please refer to the Appendix for the full ablation results. 

\begin{table}[t]
\centering
\caption{Effect of the change in the magnitude of $\gamma$. Baseline refers to the AdvRush with default $\gamma$ of 0.01. The best result in each column is in bold, and the second best result is underlined.}
\setlength{\tabcolsep}{4pt}
\vspace{5pt}
\renewcommand{\arraystretch}{1.1}
\begin{tabular}{l|cccc}
\toprule
$\gamma$ & Clean & FGSM & PGD$^{20}$ & PGD$^{100}$ \\
\bottomrule\toprule
0.001 (x 0.1) & 85.65\% & 60.04\% & 52.70\% & 52.39\% \\
0.005 (x 0.5) & \underline{85.68\%} & \underline{60.31\%} & 52.93\% & 52.61\% \\
0.01 (baseline) & \textbf{87.30\%} & \textbf{60.87\%} & 53.07\% & \underline{52.80\%} \\
0.02 (x 2) & 83.15\% & 59.34\% & \underline{53.42\%} & \textbf{53.19\%} \\
0.1 (x 10) & 83.03\% & 59.69\% & \textbf{53.67\%} & 52.20\% \\
\bottomrule
\end{tabular}
\vspace{-10pt}
\label{table:gamma}
\end{table}

\subsection{Comparison against Supernet Adv. Training}
Introducing a curve regularizer to the update rule of architectural parameters can be considered as being analogous to adversarial training of architectural parameters.
Therefore, we compare AdvRush against adversarial training of the architectural parameters using two adversarial losses: 7-step PGD and FGSM.
Since adversarial training can be introduced with or without warming up $\omega$ and $\alpha$ of the supernet, we test both scenarios.
The search results can be found in~\tablename~\ref{table:diff_adv_loss}.
It appears that neither one of the adversarial losses is effective at inducing additional robustness in the searched architecture.
We conjecture that the inner and the outer objectives of the bi-level optimization collide with each other when trying to fit clean and perturbed data alternatively, thereby disrupting the search process. 
In the following section, we show in detail why the architectures searched through adversarial training of the supernet are significantly less robust than the family of AdvRush architectures.
The failure of adversarial losses upholds the particular adequacy of the curve regularizer in AdvRush for discovering a robust neural architecture.

\begin{table}[t]
\centering
\caption{AdvRush compared to various standard adversarial training methods. $E_{\tiny \textrm{Warm}}$ and $E_{\tiny \textrm{Adv}}$ denote the number of epochs with and without the adversarial loss term, respectively.}
\setlength{\tabcolsep}{3pt}
\vspace{5pt}
\renewcommand{\arraystretch}{1.0}
\begin{tabular}{ccc|cccc}
\toprule
Search & $E_{\tiny \textrm{Warm}}$ & $E_{\tiny \textrm{Adv}}$ & Clean & FGSM & PGD$^{20}$ & PGD$^{100}$ \\
\bottomrule\toprule
\multirow{3}{*}{FGSM} & 0  & 50 & 83.04\% & 56.30\% & 48.76\% & 48.47\% \\ 
 & 0  & 60 & 82.82\% & 55.55\% & 48.41\% & 48.04\% \\ 
 & 50 & 10 & 82.78\% & 54.17\% & 46.48\% & 46.03\% \\ 
\midrule
\multirow{3}{*}{PGD} & 0 & 50 & 81.13\% & 53.59\% & 46.37\% & 45.95\% \\ 
 & 0  & 60 & 81.87\% & 53.46\% & 46.22\% & 45.82\% \\ 
 & 50  & 10 & 82.26\% & 54.87\% & 46.81\% & 46.63\% \\ 
\midrule
\multicolumn{3}{c|}{AdvRush} & \textbf{87.30\%} & \textbf{60.87\%} & \textbf{53.07\%} & \textbf{52.80\%} \\
\bottomrule
\end{tabular}
\vspace{-10pt}
\label{table:diff_adv_loss}
\end{table}

\begin{figure*}[t]
\begin{center}
   \includegraphics[width=\linewidth]{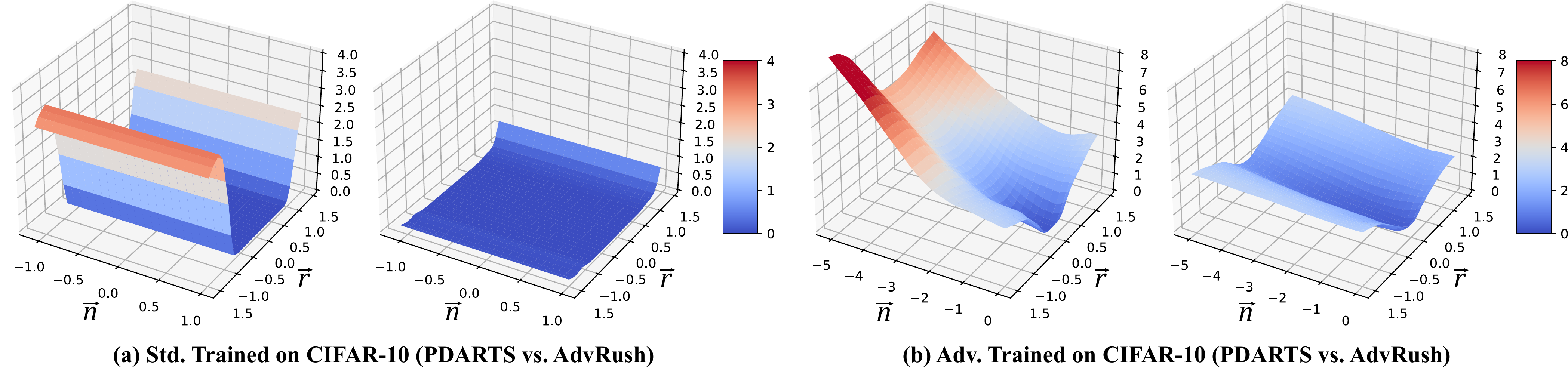}
\end{center}
   \caption{Loss landscape visualization for (a) standard-trained architectures and (b) adversarially-trained architectures. $\Vec{n}$ and $\Vec{r}$ denote perturbations in normal and random directions, respectively. The loss landscapes of AdvRush are visibly smoother than those of PDARTS.}
\label{fig:loss_land}
    \vspace{-5pt}
\end{figure*}

\begin{table*}[t]
\centering
\caption{Architecture analysis. \{M.P.; A.P.; S.; S.C.; D.C.\} each refers to max pooling, average pooling, skip connect, seperable convolution, and dilated convolution. The width and the depth of an architecture are measured following Shu \textit{et al.}~\cite{shu2019understanding}, and {N, R} denote the normal and the reduction cell. Std. Tr. and Adv. Tr. refer to the HRS score of an architecture after standard and adversarial training.}
\setlength{\tabcolsep}{7pt}
\vspace{3pt}
\renewcommand{\arraystretch}{1.0}
\begin{tabular}{c|c|c|c|c|c|c|c}
\toprule
\multirow{2}{*}{Search} & \multirow{2}{*}{Arch} & \multirow{2}{*}{Params} & \# of Operations & Width & Depth & \multicolumn{2}{c}{HRS ($\uparrow$)} \\
& & & \{M.P.; A.P.; S.; S.C.; D.C.\} & \{\textit{N, R}\} & \{\textit{N, R}\} & Std. Tr. & Adv. Tr.\\
\bottomrule\toprule
\multicolumn{2}{c|}{PDARTS} & 3.4M & \{0; 1; 3; 9; 2\} & \{2.5c, 3c\} & \{4, 3\} & 69.92 & 64.10\\
\midrule
\multirow{5}{*}{AdvRush} & Arch 0 & 4.2M & \{0; 2; 4; 9; 1\} & \{4c; 3c\} & \{2, 3\} & 71.09 & 66.01 \\
& Arch 1 & 4.2M & \{0; 0; 5; 8; 3\} & \{4c, 3c\} & \{2, 3\} & 63.50 & 65.25\\
& Arch 2 & 4.2M & \{0; 0; 4; 10; 2\} & \{3.5c, 3.5c\} & \{3, 3\} & 71.06 & 65.44 \\
& Arch 3 & 3.8M & \{0; 0; 6; 7; 3\} & \{4c, 3c\} & \{2, 3\} & 64.24 & 64.05 \\
& Arch 4 & 3.7M & \{0; 0; 5; 8; 3\} & \{4c, 4c\} & \{2, 2\} & 66.27 & 65.20 \\
\midrule
& Arch 5 & 2.9M & \{4; 0; 3; 2; 7\} & \{4c, 3.5c\} & \{2, 3\} & 61.57 & 61.44 \\
& Arch 6 & 3.0M & \{4; 0; 3; 1; 8\} & \{4c, 3c\} & \{2, 3\} & 67.36 & 61.10 \\
Supernet & Arch 7 & 2.3M & \{5; 1; 6; 1; 3\} & \{4c, 3c\} & \{2, 3\} & 44.55 & 59.53 \\
Adv. Tr. & Arch 8 & 2.3M & \{1; 0; 4; 0; 11\} & \{3c, 3c\} & \{4, 3\} & 38.87 & 59.01 \\
& Arch 9 & 2.1M & \{1; 0; 6; 0; 9\} & \{3.5c, 3c\} & \{4, 3\} & 34.52 & 59.08 \\
& Arch 10 & 2.3M & \{0; 5; 7; 1; 3\} & \{4c, 2.5c\} & \{2, 3\} & 38.15 & 59.66 \\
\bottomrule
\end{tabular}
\vspace{-10pt}
\label{table:arch_compa}
\end{table*}

\subsection{Examination of the Searched Architecture}
To show that the AdvRush architecture in fact has a relatively smooth input loss landscape, we compare the input loss landscapes of the AdvRush and the PDARTS architectures after standard training and adversarial training in~\figurename~\ref{fig:loss_land}.
The loss landscapes are visualized using the same technique as utilized by Moosavi \textit{et al.}~\cite{moosavi2019robustness}.
Degrees of perturbation in input data for standard-trained and adversarially-trained architectures are set differently to account for the discrepancy in their sensitivity to perturbation.
Independent of the training method, the input loss landscape of the AdvRush architecture is visibly smoother.
The visualization results provide strong empirical support for AdvRush by demonstrating that the search result is aligned with our theoretical motivation.

We analyze the architectures used in our experiments to provide a meaningful insight into what makes an architecture robust.
To begin with, we find that architectures that are based on the DARTS search space are generally more robust than hand-crafted architectures.
The DARTS search space is inherently designed to yield an architecture with dense connectivity and complicated feature reuse.
We believe that the complex wiring pattern of the DARTS search space allows derived architectures to be more robust than others, as observed by Guo \textit{et al.}~\cite{guo2020meets}.

Furthermore, we compare the twelve architectures searched from the DARTS search space, through Harmonic Robustness Score (HRS)~\cite{devaguptapu2020empirical}, a recently-introduced metric for measuring the trade-off between standard and robust accuracies.
Details regarding the calculation of HRS can be found in the Appendix.
In~\tablename~\ref{table:arch_compa}, we report the HRS of each architecture, along with the summary of its architectural details.
Please refer to the Appendix for the visualization of each architecture.
In general, robust architectures with high HRS have fewer parameter-free operations (\ie pooling and skip connect) and more separable convolution operations.
As a result, they tend to have more parameters than non-robust ones; this result coincides with the observations in Madry \textit{et al.}~\cite{madry2017towards} and Su \textit{et al.}~\cite{su2018robustness}.

Also, the fact that Arch 5 \& 6, in spite of having a fewer number of parameters, have comparable HRS to some of the larger architectures leads us to believe that the diversification of operations contributes to improving the robustness.
The operational diversity is once again observed in PDARTS and Arch 0, both of which have high HRS.
Lastly, no clear relationship between the width and the depth of an architecture and its robustness can be found, indicating that these two factors may have less influence over robustness.

\section{Conclusion}
In this work, AdvRush, a novel adversarial robustness-aware NAS algorithm, is proposed. 
The objective function of AdvRush is designed to prefer a candidate architecture with a smooth input loss landscape.
The theoretical motivation behind our approach is validated by strong empirical results. 
Possible future works include the study of a robust neural architecture for multimodal datasets and expansion of the search space to include more diversified operations.

\clearpage

{
\small
\balance
\bibliographystyle{ieee_fullname}
\bibliography{main}
}

\clearpage

\section*{Appendices}
\setcounter{section}{0}
\renewcommand\thesection{A\arabic{section}}
\setcounter{table}{0}
\renewcommand{\thetable}{A\arabic{table}}
\setcounter{figure}{0}
\renewcommand{\thefigure}{A\arabic{figure}}
\setcounter{equation}{0}
\renewcommand{\theequation}{A\arabic{equation}}

\section{AdvRush Implementation Details}
Following DARTS, the search space of AdvRush includes following operations:
\begin{multicols}{2}
\begin{itemize}
\setlength\itemsep{0.1em}
    \item Zero Operation
    \item Skip Connect
    \item 3x3 Average Pooling
    \item 3x3 Max Pooling
    \item 3x3 Separable Conv
    \item 5x5 Separable Conv
    \item 3x3 Dilated Conv
    \item 5x5 Dilated Conv
\end{itemize}
\end{multicols}
The pseudo code is presented in Algorithm 1.
\begin{algorithm}[h]
\SetAlCapFnt{\small}
\DontPrintSemicolon
\SetKwInOut{Input}{Input}
\SetKwInOut{Output}{Output}
\Input{
$E$ = total number of epochs for search \\
$E_\mathrm{warmup}$ = number of epochs to warm-up \\
$\gamma$ = regularization strength
}
\Output{
$f^*_A(\cdot)$ = final architecture 
}
\BlankLine
Initialize $f_\mathrm{super}(\omega_0, \alpha_0)$ \;
\textbf{For} $i$ = from $1$ to $E$: \;
~~~\textbf{If} $i \leq E_\mathrm{warmup} $ \;
\small~~~~~~ Update $\omega_i$ using $\nabla_{\omega}\mathcal{L}_{\mathrm{train}}(\omega_{i-1}, \alpha_{i-1})$ (SGD) \;
~~~~~~ Update $\alpha_i$ using $\nabla_{\alpha}\mathcal{L}_{\mathrm{val}}(\omega_{i}, \alpha_{i-1})$ (Adam)\;

\normalsize~~~\textbf{Else} \;
\small~~~~~~ Update $\omega_i$ using $\nabla_{\omega}\mathcal{L}_{\mathrm{train}}(\omega_{i-1}, \alpha_{i-1})$ (SGD) \;
~~~~~~ Update $\alpha_i$ using $\nabla_{\alpha}[\mathcal{L}_{\mathrm{val}}(\omega_{i}, \alpha_{i-1}) + \gamma\mathcal{L}_\lambda]$ (Adam)\;

\normalsize~~~\textbf{End}

\textbf{End}

Derive $f^*_A(\cdot)$ through discretization rule of DARTS from $f_\mathrm{super}(\omega_E, \alpha_E)$

\caption{AdvRush (Search)}
\label{alg:advrush}
\end{algorithm}

\section{Hyperparameters and Datasets}
For adversarial training, a different set of hyperparameters for each dataset.
Hyperparameter settings for the datasets used in our experiments are provided in~\tablename~\ref{table:hyperparam}.
Details regarding the datasets are provided in~\tablename~\ref{table:dataset_config}. 
\begin{table}[h]
\centering
\caption{Details of hyperpameters used for adversarial training on different datasets. Learning rate is decayed by the factor of 0.1 at selected epochs. CIFAR refers to both CIFAR-10 and -100. }
\setlength{\tabcolsep}{4.5pt}
\vspace{3pt}
\renewcommand{\arraystretch}{0.9}
\begin{tabular}{l|ccc}
\toprule
& CIFAR & SVHN & Tiny-ImageNet \\
\bottomrule \toprule
Optimizer & SGD & SGD & SGD \\
\midrule
Momentum & 0.9 & 0.9 & 0.9 \\
\midrule
Weight decay & 1e-4 & 1e-4 & 1e-4 \\
\midrule
Epochs & 200 & 200 & 90 \\
\midrule
LR & 0.1 & 0.01 & 0.1 \\  
\midrule
LR decay & (100, 150) & (100, 150) & (30, 60) \\
\bottomrule
\end{tabular}
\vspace{-10pt}
\label{table:hyperparam}
\end{table}

\section{Difference in $\mathcal{L}_\lambda$}
\begin{figure}[h]
    \centering
    \includegraphics[width=0.9\linewidth]{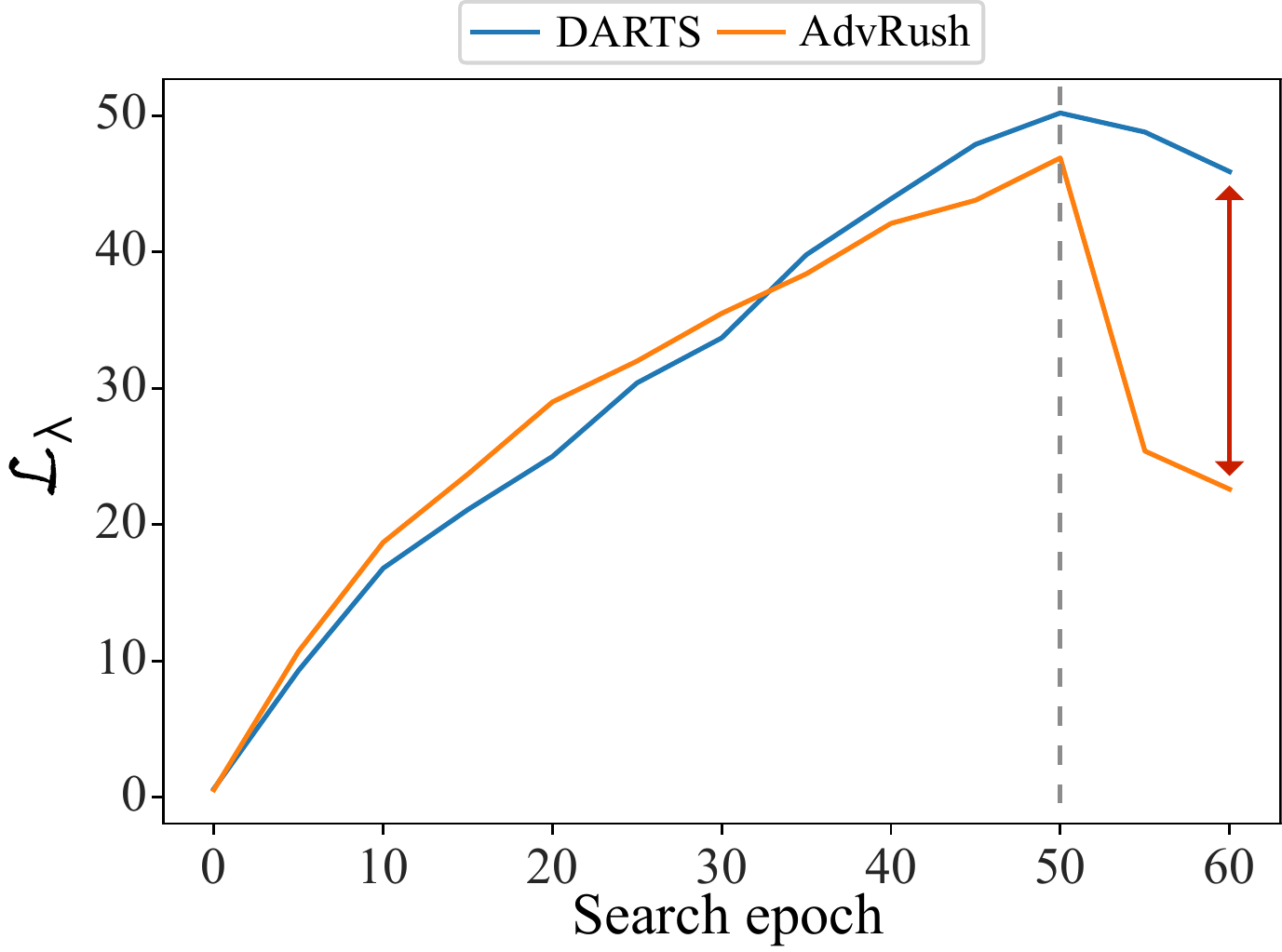}
    \caption{The difference in value of $\mathcal{L}_\lambda$ between DARTS and AdvRush. $\mathcal{L}_\lambda$ is introduced at Epoch $50$, noted in a gray dotted line. The red arrow points out the gap in $\mathcal{L}_\lambda$ between the two algorithms, caused by different $\alpha$ update rules.}
    \label{fig:darts_advrush}
\end{figure}

In~\figurename~\ref{fig:darts_advrush}, we visualize how the difference in $\alpha$ update rule between DARTS and AdvRush affects the value of $\mathcal{L}_\lambda$.
We use $\gamma$ of $0.01$ for AdvRush, and $\mathcal{L}_\lambda$ is introduced at the $50^{th}$ epoch.
Notice that AdvRush experiences a steeper drop in $\mathcal{L}_\lambda$; this phenomenon implies that the final supernet of AdvRush is indeed smoother than that of DARTS.
Since the only difference in the two compared search algorithms is the learning objective of $\alpha$, it is safe to assume that this extra smoothness is induced solely by the difference in $\alpha$. 

\section{More Input Loss Landscapes}
\begin{figure*}[t]
    \centering
    \includegraphics[width=\linewidth]{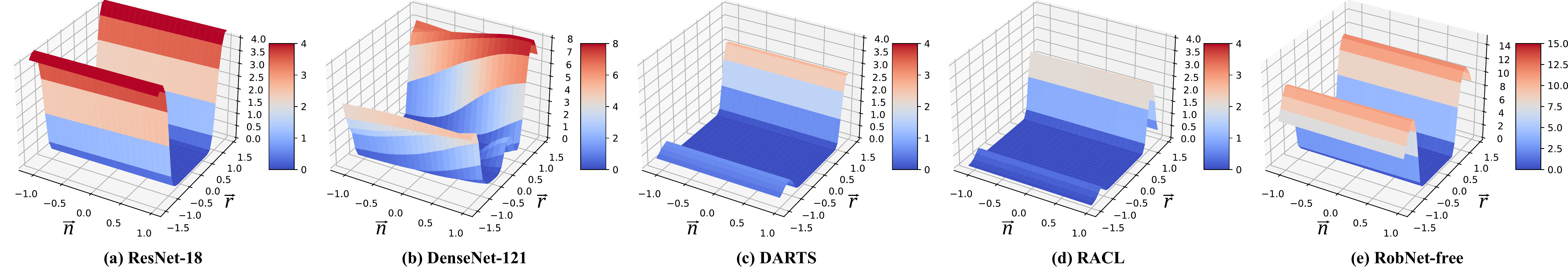}
    \caption{Input loss landscapes after standard training on CIFAR-10.}
    \label{fig:appen_std_land}
    \vspace{-5pt}
\end{figure*}

\begin{figure*}[t]
    \centering
    \includegraphics[width=\linewidth]{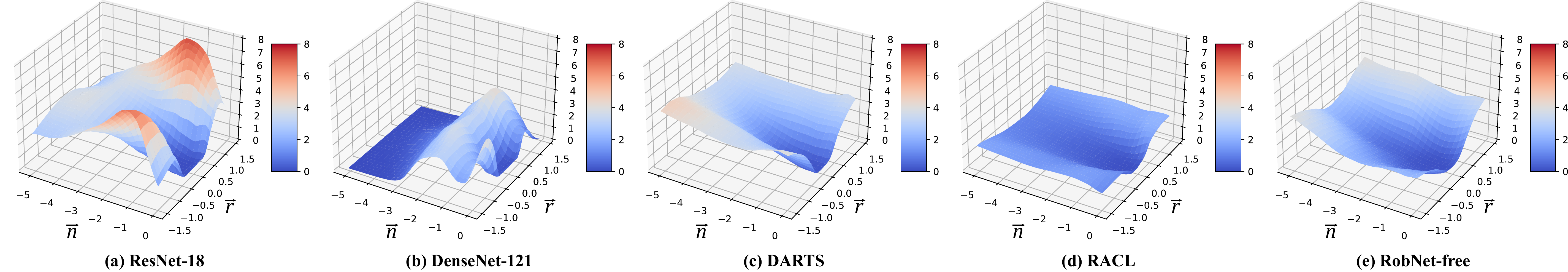}
    \caption{Input loss landscapes after adversarial training on CIFAR-10.}
    \label{fig:appen_adv_land}
    \vspace{-25pt}
\end{figure*}

We provide additional input loss landscapes of other tested architectures after standard training (\figurename~\ref{fig:appen_std_land}) and adversarial training (\figurename~\ref{fig:appen_adv_land}).
All loss landscapes are drawn using the CIFAR-10 dataset.
In both figures, $\Vec{n}$ and $\Vec{r}$ denote perturbations in normal and random directions, respectively.
Degrees of perturbation in input data for standard trained and adversarially trained architectures are set differently to account for the discrepancy in their sensitivity to perturbation.

\section{Architecture Visualization}
\begin{figure*}[t]
    \centering
    \includegraphics[width=\linewidth]{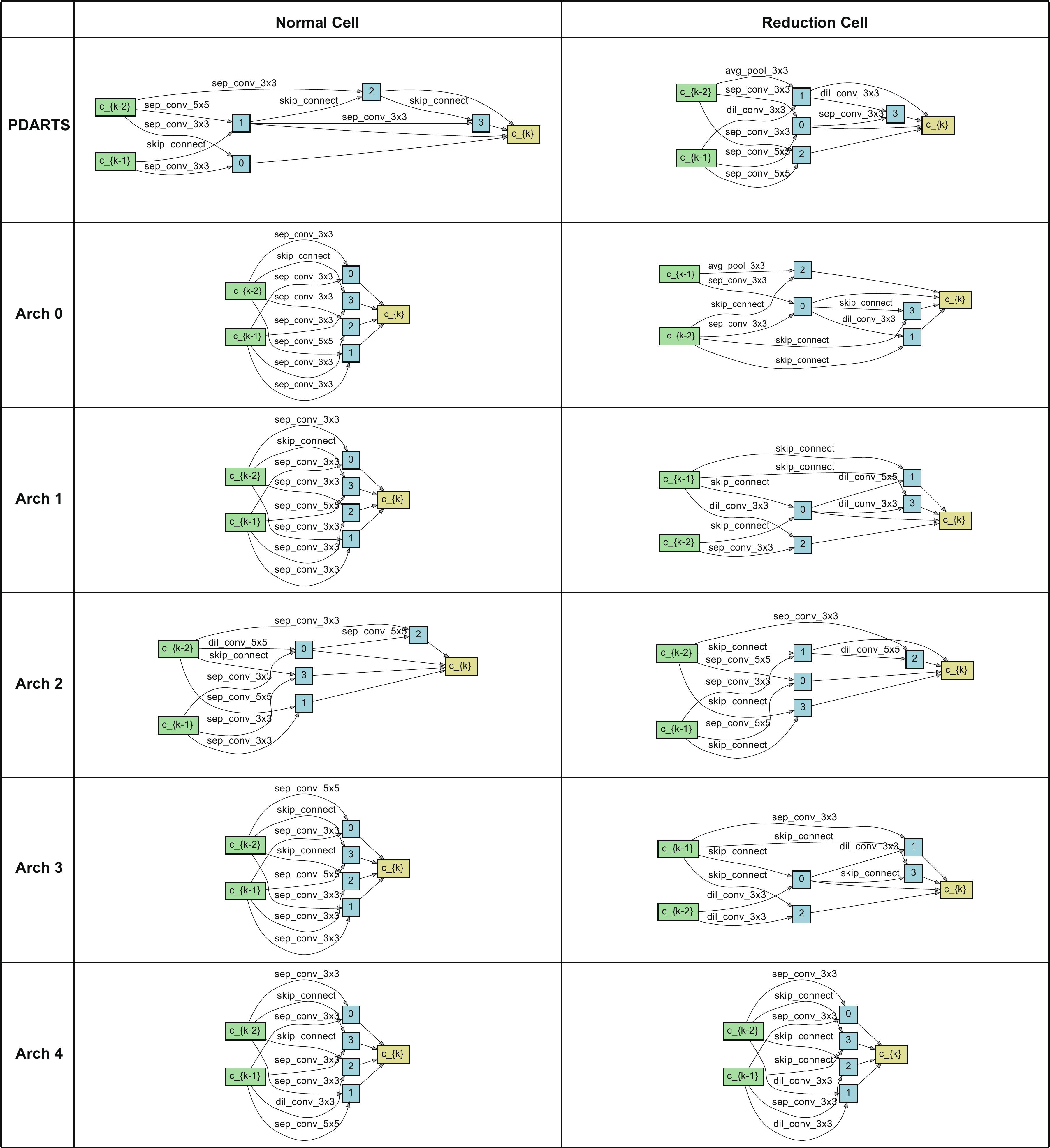}
    \caption{Architectures searched by PDARTS and AdvRush. A normal cell maintains the dimension of input feature maps, while a reduction cell reduces it. The entire neural network is constructed by stacking 18 normal cells and 2 reduction cells following the DARTS convention. The reduction cell is placed at the $1/3$ and $2/3$ points of the network.}
    \label{fig:cell_table1}
    \vspace{-5pt}
\end{figure*}

\begin{figure*}[t]
    \centering
    \includegraphics[width=\linewidth]{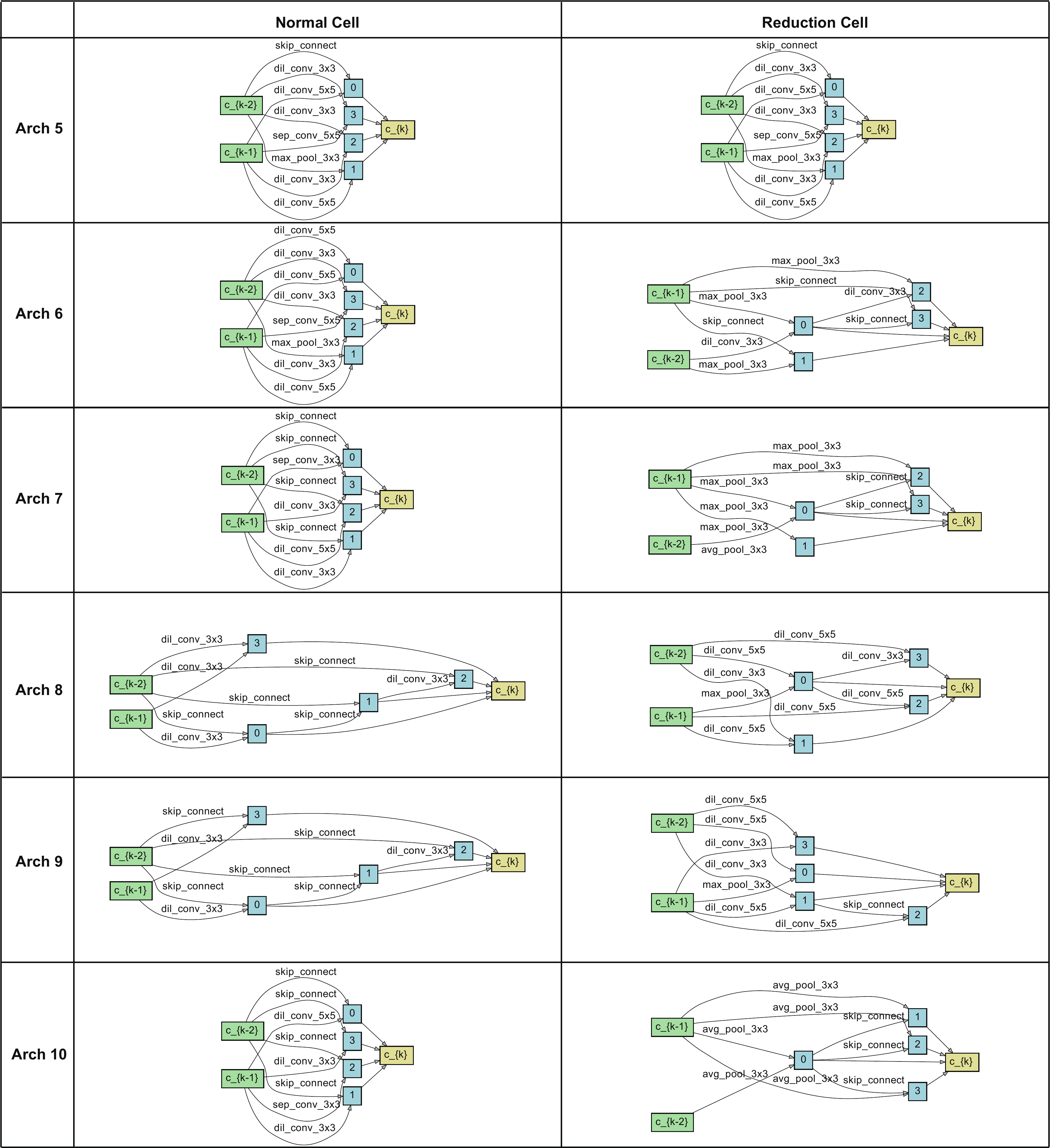}
    \caption{Architectures searched by adversarial training of the supernet. A normal cell maintains the dimension of input feature maps, while a reduction cell reduces it. The entire neural network is constructed by stacking 18 normal cells and 2 reduction cells following the DARTS convention. The reduction cell is placed at the $1/3$ and $2/3$ points of the network.}
    \label{fig:cell_table2}
    \vspace{-5pt}
\end{figure*}

The architectures used for the analysis in Section \textcolor{red}{6.3} of the main text are visualized in~\figurename~\ref{fig:cell_table1} and~\ref{fig:cell_table2}.

\section{HRS Calculation}
HRS is calculated as follows:
\begin{equation}
    HRS = \frac{2CR}{C+R}
\end{equation}
where C denotes the clean accuracy, and R denotes the robust accuracy.
For standard-trained architectures, we use the robust accuracy under FGSM attack for R because the robust accuracy of standard-trained architectures under PGD attack reaches near-zero.
For adversarially-trained architectures, we use the robust accuracy under PGD$^{20}$ attack for R.
C, R, and HRS values for all architectures used in Section \textcolor{red}{6.3} of the main text can be found in~\tablename~\ref{table:more_arch_result}.

\section{Extended Results on Other Datasets}
\begin{figure*}[t]
    \centering
    \includegraphics[width=\linewidth]{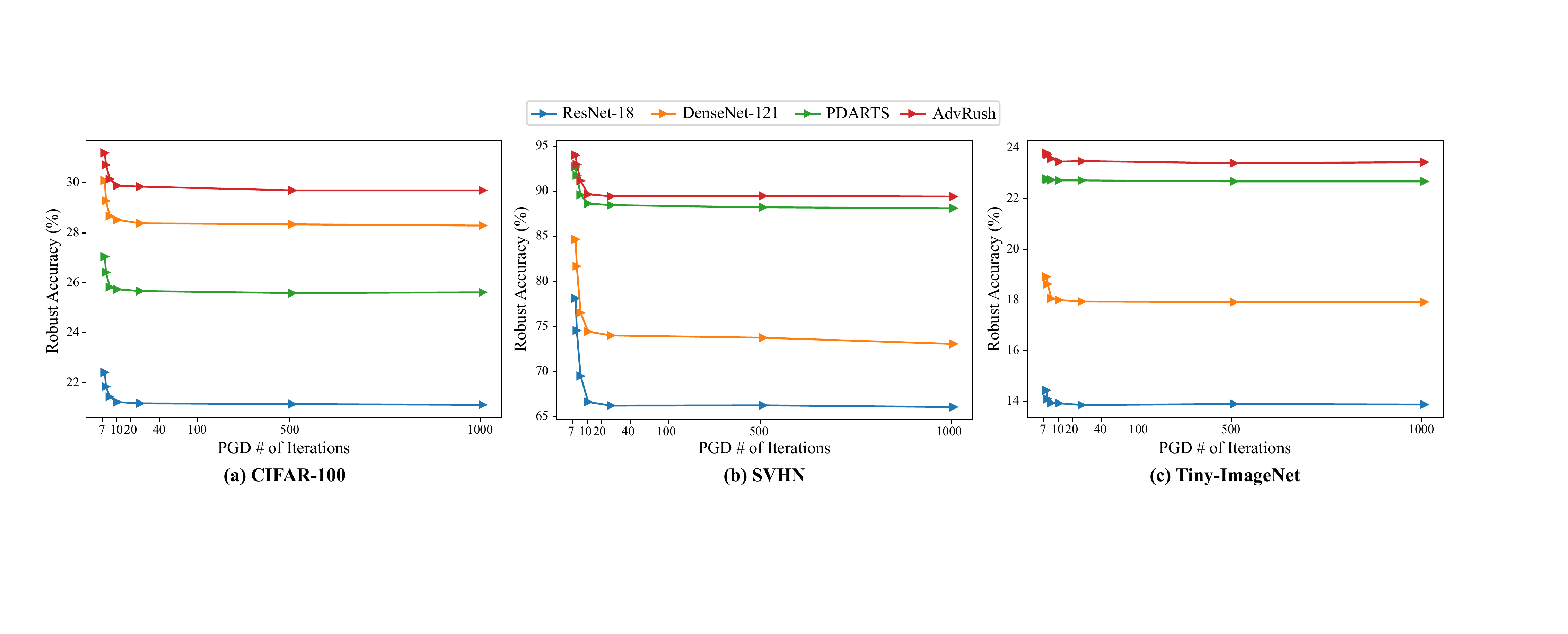}
    \vspace{-57pt}
    \caption{Robust accuracies on (a) CIFAR-100, (b) SVHN, and (c) Tiny-ImageNet under different iterations of PGD attack.}
    \label{fig:appen_alldataset}
\end{figure*}

In addition to the results in the main text, we conduct PGD attack with different number of iterations and evaluate the robust accuracy on CIFAR-100, SVHN, and Tiny-ImageNet. 
We use varying iterations from $7$ to $1000$.
The results are visualized in~\figurename~\ref{fig:appen_alldataset}.
AdvRush consistently outperforms other architectures, regardless of the strength of the adversary.

\section{Full Ablation Results}
In~\tablename~\ref{table:gamma_full}, the full ablation analysis of AdvRush including the robust accuracies under AutoAttack is presented.
No matter the value of $\gamma$ used, architectures searched with AdvRush achieve high robust accuracies under AutoAttack.
Such results suggest that AdvRush does not require excessive tuning of $\gamma$ to search for a robust neural architecture.

\begin{table}[h]
\centering
\caption{Effect of the change in the magnitude of $\gamma$. Baseline refers to the AdvRush with default $\gamma$ of 0.01. The best result in each column is in bold, and the second best result is underlined.}
\setlength{\tabcolsep}{3pt}
\vspace{5pt}
\renewcommand{\arraystretch}{1.2}
\begin{tabular}{l|ccc|ccc}
\toprule
\multirow{2}{*}{Arch} & \multicolumn{3}{c|}{Std. Tr.} & \multicolumn{3}{c}{Adv. Tr.} \\
& Clean & FGSM & HRS & Clean & PGD$^{20}$ & HRS \\
\bottomrule\toprule
PDARTS & 97.49\% & 54.51\% & 69.92 & 85.37\% & 51.32\% & 64.10 \\
Arch 0 & 97.58\% & 55.91\% & 71.09 & 87.30\% & 53.07\% & 66.01 \\
Arch 1 & 95.59\% & 47.54\% & 63.50 & 85.65\% & 52.70\% & 65.25 \\
Arch 2 & 95.55\% & 56.57\% & 71.06 & 85.68\% & 52.93\% & 65.44 \\
Arch 3 & 95.24\% & 48.46\% & 64.24 & 83.15\% & 53.42\% & 64.05 \\
Arch 4 & 95.45\% & 50.75\% & 66.27 & 83.03\% & 53.67\% & 65.20 \\
Arch 5 & 95.05\% & 45.53\% & 61.57 & 83.04\% & 48.76\% & 61.44 \\
Arch 6 & 94.93\% & 52.20\% & 67.36 & 82.82\% & 48.41\% & 61.10 \\
Arch 7 & 94.01\% & 29.19\% & 44.55 & 82.78\% & 46.48\% & 59.53 \\
Arch 8 & 93.24\% & 24.55\% & 38.87 & 81.13\% & 46.37\% & 59.01 \\
Arch 9 & 93.08\% & 21.19\% & 34.52 & 81.87\% & 46.22\% & 59.08 \\
Arch 10 & 93.88\% & 23.94\% & 38.15 & 82.26\% & 46.81\% & 59.66 \\
\bottomrule
\end{tabular}
\vspace{-10pt}
\label{table:more_arch_result}
\end{table}

\begin{table*}[t]
\centering
\caption{Effect of the change in the magnitude of $\gamma$. Baseline refers to AdvRush with default $\gamma$ of 0.01. The best result in each column is in bold, and the second best result is underlined.}
\vspace{5pt}
\renewcommand{\arraystretch}{1.1}
\begin{tabular}{l|ccccccccc}
\toprule
$\gamma$\text{-value} & Clean & FGSM & PGD$^{20}$ & PGD$^{100}$ & APGD\textsubscript{CE} & APGD\textsuperscript{T} & FAB\textsuperscript{T} & Square & AA \\
\bottomrule\toprule
0.001 (x 0.1) & 85.65\% & 60.04\% & 52.70\% & 52.39\% & 51.39\% & 49.16\% & 49.13\% & 49.13\% & 49.13\% \\
0.005 (x 0.5) & \underline{85.68\%} & \underline{60.31\%} & 52.93\% & 52.61\% & \underline{51.77\%} & \underline{49.63\%} & \underline{49.62\%} & \underline{49.62\%} & \underline{49.62\%} \\
0.01 (baseline) & \textbf{87.30\%} & \textbf{60.87\%} & 53.07\% & \underline{52.80\%} & 50.05\% & \textbf{50.04\%} & \textbf{50.04\%} & \textbf{50.04\%} & \textbf{50.04\%} \\
0.02 (x 2) & 83.15\% & 59.34\% & \underline{53.42\%} & \textbf{53.19\%} & \textbf{52.22\%} & 49.60\% & 49.60\% & 49.60\% & 49.60\% \\
0.1 (x 10) & 83.03\% & 59.69\% & \textbf{53.67\%} & 52.20\% & 51.22\% & 49.12\% & 49.11\% & 49.11\% & 49.11\% \\
\bottomrule
\end{tabular}
\label{table:gamma_full}
\end{table*}

\begin{table*}[h]
\centering
\caption{Summary of datasets used for the experiments. CIFAR-10, CIFAR-100, and SVHN, which are available directly from torchvision, are split only into train and test sets by default.}
\vspace{5pt}
\setlength{\tabcolsep}{4pt}
\renewcommand{\arraystretch}{1.1}
\begin{tabular}{l|ccccc}
\toprule
Dataset & \# of Train Data & \# of Validation Data & \# of Test Data & \# of Classes & Image Size \\
\bottomrule\toprule
CIFAR-10 & 50,000 & - & 10,000 & 10 & $(32 \times 32)$ \\
CIFAR-100 & 50,000 & - & 10,000 & 100 & $(32 \times 32)$ \\
SVHN & 73,257 & - & 26,032 & 10 & $(32 \times 32)$ \\
Tiny-ImageNet & 100,000 & 5,000 & 5,000 & 200 & $(64 \times 64)$ \\
\bottomrule
\end{tabular}
\label{table:dataset_config}
\end{table*}

\end{document}